\documentclass[sigplan,screen,twocolumn,nonacm]{acmart}

\usepackage{microtype}
\usepackage[normalem]{ulem}
\usepackage{xspace}
\usepackage{relsize}
\usepackage{soul}
\usepackage{makecell}
\usepackage{hyphenat}
\usepackage{multirow}
\usepackage{booktabs}
\usepackage{amssymb}
\usepackage{amsmath} 
\usepackage{dblfloatfix}
\usepackage{float}
\usepackage{tabularx}
\usepackage{booktabs}
\usepackage{seqsplit}
\usepackage{float}

\newcommand{\good}{\checkmark}
\newcommand{\bad}{\text{\sffamily X}}

\usepackage{physics2}\usephysicsmodule{ab}
\usepackage{enumitem}

\usepackage{comment}

\usepackage{xcolor}
\usepackage{graphicx} %
\usepackage{import}
\usepackage{svg}
\usepackage{placeins}
\usepackage{tikz}\usetikzlibrary{arrows.meta,positioning,decorations}
\usepackage{pgfplots}\pgfplotsset{compat=1.18}
\usepackage{graphicx}
\usepackage{capt-of}  %
\usepackage{cuted}    %

\usepackage{algorithm}
\usepackage{algpseudocode}

\usepackage{booktabs}
\usepackage{threeparttable}

\usepackage{titlesec}

\titleformat{\subsubsection}{\normalfont\normalsize\bfseries}{\thesubsubsection}{1em}{}

\usepackage[group-separator={,}]{siunitx}

\newcommand{\Cpp}{C\texttt{++}\xspace}

\newcommand{\mysubsubsection}[1]{{\vspace{0.25em}\noindent\ul{\textbf{\textit{#1.\xspace}}}}}

\renewcommand\footnotetextcopyrightpermission[1]{}
\begin{document}

\title{SlimPack: Fine-Grained Asymmetric Packing for Balanced and Efficient Variable-Length LLM Training}
\date{May 2025}

%
\author{Yuliang Liu}
\authornote{Equal contribution.}
\authornote{Corresponding author.}
\email{liuyuliang@kuaishou.com}
\affiliation{%
\institution{Kling Infra, Kuaishou Technology}
\country{}
}

\author{Guohao Wu}
\authornotemark[1]
\email{wuguohao03@kuaishou.com}
\affiliation{%
\institution{Kling Infra, Kuaishou Technology}
\country{}
}

\author{Shenglong Zhang}
\authornotemark[1]
\email{zhangshenglong@kuaishou.com}
\affiliation{%
\institution{Kling Infra, Kuaishou Technology}
\country{}
}

\author{Wei Zhang}
\email{zhangwei45@kuaishou.com}
\affiliation{%
\institution{Kling Infra, Kuaishou Technology}
\country{}
}

\author{Qianchao Zhu}
\email{zhuqianchao@kuaishou.com}
\affiliation{%
\institution{Kling Infra, Kuaishou Technology}
\country{}
}

\author{Zhouyang Li}
\email{lizhouyang@kuaishou.com}
\affiliation{%
\institution{Kling Infra, Kuaishou Technology}
\country{}
}

\author{Chenyu Wang}
\email{wangchenyu05@kuaishou.com}
\affiliation{
\institution{Kling Infra, Kuaishou Technology}
\country{}
}

\begin{abstract}
The efficient distributed training of Large Language Models (LLMs) is severely hampered by the extreme variance in context lengths. This data heterogeneity, amplified by conventional packing strategies and asymmetric forward-backward costs, leads to critical inefficiencies such as cascading workload imbalances and severe hardware underutilization. Existing solutions attempt to mitigate these challenges, but often at the expense of memory or communication efficiency.

To address these challenges, we introduce SlimPack, a framework that fundamentally rethinks data packing and scheduling by decomposing samples into fine-grained slices. This slice-level decomposition immediately mitigates critical memory and communication bottlenecks by transforming large, volatile workloads into a stream of smaller, manageable units. 
This flexibility is then harnessed for our core innovation, Asymmetric Partitioning, which assembles balanced scheduling units uniquely optimized for the different demands of the forward and backward passes.
Orchestrated by a two-phase solver and a high-fidelity simulator, SlimPack holistically resolves imbalances across all parallel dimensions. Extensive experiments demonstrate that SlimPack achieves up to a $2.8\times$  training throughput improvement over baselines, breaking the conventional trade-off by delivering both superior balance and high resource efficiency.

\end{abstract}

\maketitle
\renewcommand{\shortauthors}{Liu et al.}

\section{Introduction}
\label{sec:introduction}

The relentless growth in the scale of LLMs~\cite{Gemini,Llama3,GPT-4} has been a primary driver of their remarkable capabilities, yet this expansion has simultaneously intensified the challenges of efficient distributed training. A fundamental, yet often overlooked, bottleneck is the extreme variance in sequence lengths found in real-world training corpora~\cite{LongAlign,liu2023lost}. Datasets exhibit a pronounced long-tail distribution~\cite{CommonCrawl,Github-code,wikidump}, where a small fraction of ultra-long sequences contributes a disproportionately large share of the computational workload. Existing distributed systems~\cite{MegatronSP,rasley2020deepspeed,DeepSpeedUlysses}, however, are typically designed with an assumption of workload homogeneity, employing static parallelism and scheduling strategies that are fundamentally misaligned with this data heterogeneity. This mismatch leads to severe hardware underutilization, manifesting as cascading workload imbalances across multiple parallelism dimensions.

To manage variable-length inputs, a common strategy adopted by training systems is sample-level packing~\cite{Packing, PackingAnalysis}, which typically involves combining multiple shorter sequences into a single micro-batch of fixed length.
While this reduces padding, it creates two critical challenges in hybrid parallel systems. First, due to quadratic cost of self-attention~\cite{AttentionIsAllYouNeed}, a single long-sequence sample can become a ``straggler'', dictating the execution time for an entire micro-batch. And in a hybrid system, this creates a powerful amplifier effect: a delay in one data parallelism rank propagates through the pipeline, creating a massive \textbf{Cascading Imbalance Bubble} that stalls expensive hardware across all parallel workers. Second, these strategies fail to account for the \textbf{Asymmetric Costs} of the forward and backward passes. The specialized mechanism of memory-efficient attention necessitates the re-computation of the attention score during the backward pass. This results in a forward-to-backward computational ratio of approximately $1:2.5$ for the attention mechanism, which differs from the $1:2$ ratio typical of standard GEMM operations.  Consequently, a perfectly balanced forward pass inevitably becomes imbalanced during backpropagation, reintroducing the very straggler problem the packing was meant to solve.

Existing solutions for load imbalance force a critical trade-off, typically sacrificing either memory efficiency or communication efficiency to achieve a balanced workload.
The first category, \texttt{Workload-Aware Scheduling}, seeks to balance the execution flow by strategically altering the data scheduling order. For instance, WLB-LLM~\cite{WLB-LLM} packs micro-batches to balance FLOPs and employs an outlier-delay mechanism. However, this alteration of the inherent data distribution can interfere with the statistical properties of the training process, potentially compromising model convergence.
Another category, \texttt{Parallelism Hot- Switching}, dynamically reconfigures the system's parallelism strategy or communication topology at runtime~\cite{FlexSP, ByteScale, HotSPa}. While effective at matching heterogeneous data workloads, this flexibility comes at the cost of significant \textbf{Communication Overhead}. Each reconfiguration can trigger a communication burst to reshard inputs or model parameters across different parallel layouts. This issue is particularly acute when an infrequent, ultra-long outlier forces a costly transition. Furthermore, these solutions inherit the inherent memory limitations of sample-level packing. The coarse-grained packing strategy introduces significant memory inefficiency in its pursuit of workload balance, as grouping samples of disparate lengths substantially inflates the peak activation memory footprint.

To overcome these limitations, we introduce \textbf{SlimPack}, a framework that fundamentally rethinks data packing and scheduling for variable-length LLM training. Instead of treating samples as indivisible units, SlimPack decomposes them into fine-grained slices. This flexibility allows our system to assemble perfectly balanced scheduling units, which we term \textbf{MicroPacks}. Crucially, SlimPack employs \textbf{Asymmetric Partitioning}: it creates entirely separate MicroPack configurations tailored to the distinct computational profiles of the forward and backward passes, directly solving the primary cause of pipeline imbalance.
This methodology is applied hierarchically to holistically address inefficiencies in hybrid parallelism. First, to prevent system-wide stragglers, the global batch is partitioned across Data Parallel (DP) ranks to equalize their total FLOPs. Second, within each pipeline, the asymmetrically partitioned MicroPacks eliminate internal bubbles by ensuring every stage processes a consistent workload throughout the entire execution cycle.
Furthermore, this slice-level decomposition inherently \textbf{Mitigates Memory and Communication} bottlenecks, especially for ultra-long sequences. By serializing a long sequence into a stream of smaller MicroPacks, SlimPack transforms prohibitive memory spikes into a low, stable footprint. This memory efficiency enables deeper pipelines without risking Out-Of-Memory errors and reduces the need for costly, communication-intensive context parallelism~\cite{RingAttention,DeepSpeedUlysses}.

SlimPack's complex scheduling is orchestrated by an \textbf{Two-Phase Solver} and a high-fidelity \textbf{DAG-based Simulator}. The solver first determines the optimal sample distribution across DP ranks and then formulates the slice-level packing as a Mixed-Integer Linear Program to find the ideal configuration. Crucially, it resolves the inherent imbalance between forward and backward passes, while also handling extreme outliers with a novel \textbf{DP-Merge} technique. These candidate schedules are then rigorously evaluated by the simulator. By identifying the critical path, the simulator accurately models performance—predicting throughput and revealing pipeline bubbles, to guide the selection of the optimal configuration. 

We summarize our main contributions as follows:
\begin{itemize}
    \item We identify and analyze critical bottlenecks in variable-length training: the cascading imbalance effect in hybrid parallel systems and the workload skew from asymmetric forward-backward computational costs.
    
    \item We propose SlimPack, a novel system using a fine-grained, slice-level packing paradigm. Its core is an asymmetric partitioning solver that creates balanced ``MicroPacks'' for each pass, guided by a high-fidelity DAG-based simulator to find the optimal schedule.
    
    \item We demonstrate through extensive experiments that SlimPack achieves up to a \textbf{2.8\texttimes{}} training throughput improvement over the state-of-the-art Megatron-LM framework with sample-packing,, showing superior efficiency and scalability on long context training.
\end{itemize}

\section{Background \& Challenges} 
\label{background}

Table~\ref{tab:notations} summarizes the key notations used in this paper.

\begin{table}[t]
    \centering 
    \caption{Summary of key notations.}
    \label{tab:notations}
    \begin{tabular}{@{}cc@{}} 
        \toprule
        \textbf{\quad Symbol} & \textbf{Description} \\
        \midrule
        \quad  &  \textit{\textbf{ Parallelism Dimensions}} \\
        \quad $DP$ & Data parallelism size \\
        \quad $PP$ & Pipeline parallelism size  \\
        \quad $CP$ & Context parallelism size \\
        \midrule
        \quad  &  \textit{\textbf{ Batch and Sequence Dimensions}} \\
        \quad $B$ & Global batch size (total samples per iteration) \\
        \quad $b$ & Micro-batch size (samples per fwd/bwd pass) \\
        \quad $M$ & Number of micro-batches ($M = B/b$) \\
        \quad $L$ & Sequence length of input sample \\
        \midrule
        \quad  &  \textit{\textbf{ Model Architecture}} \\
        \quad $h$ & Hidden dimension size \\
        \quad $N_l$ & Number of Transformer layers\\
        \bottomrule
    \end{tabular}
\end{table}

\subsection{Variable-Length LLM Training}

The efficiency of large-scale LLM training is fundamentally undermined by the extreme variance in sequence lengths characteristic of real-world data.  While existing distributed systems employ various strategies to manage this workload variance, they lack a holistic solution, typically forcing a trade-off that sacrifices memory or communication efficiency to achieve better load balance. This problem manifests as critical workload imbalances, which in turn create severe memory and communication bottlenecks.

\subsubsection{Sources of Workload Imbalance}
\label{sec:source_of_workload_imbalance}

\begin{figure}[h]
    \centering
    \includegraphics[width=0.9\linewidth]{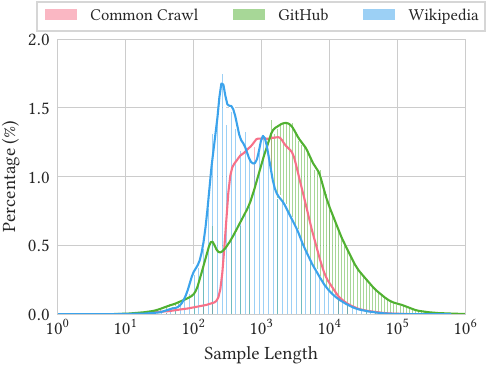}
    \caption{Distribution of sample lengths across pre-training datasets. The pronounced long-tailed pattern is a primary source of workload imbalance.}
    \label{fig:data-distribution}
\end{figure}

\noindent\textbf{Skewed Data Distributions.}
Workload imbalance in LLM training is primarily driven by the statistical nature of the input corpora. Large-scale datasets such as \emph{Common Crawl}~\cite{CommonCrawl}, \emph{GitHub}~\cite{Github-code}, and \emph{Wikipedia}~\cite{wikidump} are characterized by substantial length heterogeneity, typically following a skewed, long-tail distribution. As illustrated in Figure~\ref{fig:data-distribution}, 
While the vast majority of samples are short (e.g., nearly 80\% of samples may be shorter than 4K tokens), a disproportionately large share of the total token workload originates from a tiny fraction of extremely long sequences. For instance, it is common for ~1\% of the longest samples—often exceeding 128K tokens—to contribute nearly half of the total computational workload. This heterogeneity is particularly problematic in a distributed setting. The presence of a few outlier long sequences creates a classic straggler effect, where the overall execution time for a batch is dictated by these outliers. Consequently, parallel workers assigned shorter sequences are forced into prolonged idle states, leading to severe hardware under-utilization and a significant degradation in end-to-end training throughput.

\noindent\textbf{Quadratic Cost of Attention.} \label{sec:scaling_disparity} The workload imbalance from skewed data is dramatically amplified by the core characteristic of the Transformer: the quadratic complexity of the self-attention mechanism  $O(L^2h)$. This property is the primary architectural source of imbalance. It makes the total number of tokens in a micro-batch a poor proxy for its actual workload. While feed-forward networks scale linearly ($O(Lh^2)$), the quadratic cost of attention dominates. Consequently, a micro-batch with a single long sequence has a far greater computational cost than another of the same final length composed of multiple packed shorter sequences. This disparity is a direct cause of the straggler effect.

\subsubsection{Training with Long Context}

Training LLMs with extended context samples presents unique system-level challenges. To manage the demands, distributed training frameworks rely on a combination of parallelism strategies and specialized techniques. 

\noindent\textbf{Data Parallelism (DP).} Data Parallelism partitions the data batch across multiple devices, each holding a model replica, and synchronizes gradients via an \texttt{AllReduce} operation after local computation. This approach is memory-intensive due to the replicated model and large activation memory from long sequences. To mitigate this, sharded data parallelism techniques like ZeRO~\cite{Zero} partition model states across devices at the cost of increased communication.

\noindent\textbf{Context Parallelism (CP).} Context Parallelism~\cite{DeepSpeedUlysses, brandon2023striped, RingAttention} directly addresses the activation memory bottleneck by partitioning a single long sequence and its activations across multiple devices. This intra-data partitioning requires substantial communication, such as \texttt{Ring}-based~\cite{RingAttention} or \texttt{All-to-All}~\cite{DeepSpeedUlysses} collectives, to exchange key and value tensors for the distributed attention computation. This communication overhead is a primary performance bottleneck, especially for shorter sequences that are forced to participate.

\noindent\textbf{Pipeline Parallelism (PP).} Pipeline Parallelism~\cite{fan2021dapple, huang2019gpipe, huang2024re, li2021terapipe, harlap2018pipedream, narayanan2021memory, narayanan2021efficient} partitions a model's layers across devices (stages) and processes a batch as a stream of smaller micro-batches, requiring only minimal \texttt{Peer-to-Peer} communication. While communication-efficient, its performance is extremely sensitive to workload imbalance, as a single straggler micro-batch can create execution bubbles that propagate through all pipeline stages and stall expensive hardware.

\noindent\textbf{Hybrid Parallelism.} Modern frameworks~\cite{narayanan2021efficient, li2023colossal, DeepSpeedUlysses} combine strategies like DP, PP, and CP into a hybrid approach, often organizing devices into a multi-dimensional mesh. A key limitation of existing systems is their reliance on a static mesh like a fixed $\text{DP}\times\text{CP}$ grid, which is fundamentally misaligned with the heterogeneous workloads of variable-length data. This mismatch motivates recent work~\cite{HotSPa, FlexSP, ByteScale, WLB-LLM} on adaptive and flexible parallelism topologies.

\noindent\textbf{Activation Memory Management.} Storing activations for the backward pass is a primary memory bottleneck in long-context training. To manage this, 
activation checkpointing (or recomputation)~\cite{chen2016training, Checkmate} is a critical technique that trades computation for memory. It frees activations after the forward pass and recomputes them during the backward pass, reducing peak memory usage at the cost of increased computational overhead. Advanced systems~\cite{10.1145/3620666.3651359} can even apply this adaptively based on the workload of each micro-batch or pipeline stage. An alternative approach is activation offloading~\cite{Zero, rajbhandari2021zero, ByteScale}, which moves activations to more abundant host memory. Although constrained by limited PCIe bandwidth, the data transfer cost is often masked by the substantial computation of long sequences. 

\begin{figure*}[!tb]
  \centering
  \includegraphics[
    width=\linewidth
  ]{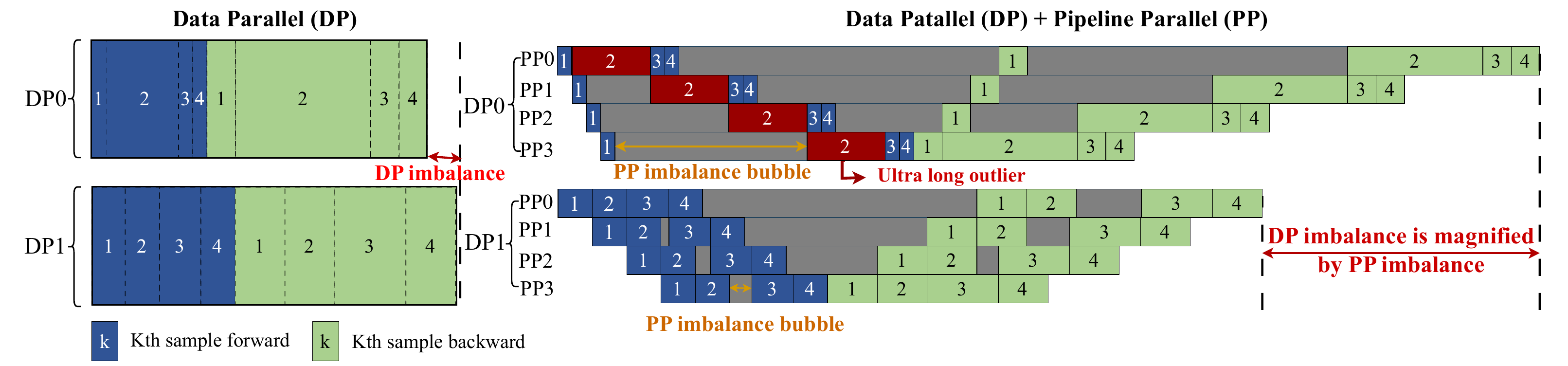}
    \caption{Illustration of the \textbf{amplifier effect} in hybrid parallelism. 
    In a pure DP system (Left), minor workload variations between workers create a small, manageable imbalance. 
    In a hybrid DP+PP system (Right), the strict micro-batch synchronization required by PP magnifies these minor delays. A single straggler (e.g., the "ultra long outlier" in DP0) forces subsequent pipeline stages to wait, creating a large \textbf{cascading imbalance bubble} and significant hardware idleness.}
    \label{fig:imba_bubble_wide}
  \label{fig:imba_bubble_wide}
\end{figure*}

\subsection{Obervation \& Challenges}

\label{sec:observations_challenges}\label{subsec:challenges}

Current training frameworks face significant challenges when handling variable-length inputs, with workload imbalance-induced device idling emerging as the primary bottleneck. This issue is severely exacerbated by the presence of extremely long outliers. While sample-level packing has been adopted as a common solution, its coarse-grained nature fails to fully resolve imbalance issues. Moreover, it introduces substantial memory pressure and communication bottlenecks. Beyond data heterogeneity, we also identify that the computational asymmetry between forward and backward passes in LLM training introduces additional pipeline bubbles, further compounding the imbalance problem.

\subsubsection{Challenges from Sample-Level Packing}
The conventional approach of treating each packed sample as an indivisible unit of work presents significant challenges in cascading imbalance bubbles, memory consumption, and communication overhead.

\noindent\textbf{The Amplifier Effect of Cascading Imbalance.}
While a pure Data Parallelism system can tolerate some workload variance between workers, the strict, fine-grained synchronization required by Pipeline Parallelism creates a powerful amplifier effect. As illustrated in Figure~\ref{fig:imba_bubble_wide}, the presence of an ultra-long outlier sample induces a substantial processing delay. This delay propagates downstream through the pipeline, starving subsequent stages of input and culminating in a severe pipeline imbalance bubble. Furthermore, such a straggler micro-batch does not only stall its own Data Parallel group. It forces all peer devices to idle during the gradient synchronization phase, waiting for the straggler to complete its computation. This mandatory synchronization induces severe resource underutilization, thereby amplifying the performance penalty across the entire system. Coarse-grained sample-level packing lacks an efficient mechanism to partition extreme outliers without resorting to costly cross-machine communication. The resulting localized pipeline bubbles then propagate globally via the aforementioned amplifier effect, causing a dramatic drop in end-to-end throughput.

\noindent\textbf{Memory Pressure and Communication Bottlenecks.}
This coarse granularity also exacerbates memory and communication challenges. A significant memory trade-off stems from the pipeline parallelism requirement to maintain $p$ computational units concurrently in flight (where $p$ denotes the pipeline parallelism size), which directly determines the peak memory footprint. As sample-level packing inherently constructs larger composite samples, it leads to substantial memory expansion. The situation becomes critical with ultra-long outliers: matching the outlier's length strains memory, but matching its computational FLOPs under quadratic-attention complexity can create prohibitively long, memory-catastrophic packs. 
Furthermore, the extreme memory footprint may necessitate the use of sequence parallelism to process a single outlier. This can force the corresponding parallel group to span multiple nodes, thereby triggering substantial and inefficient cross-node communication, which further degrades overall training throughput.
\begin{figure}[h!]
    \centering
      \includegraphics[
    width=\linewidth
  ]{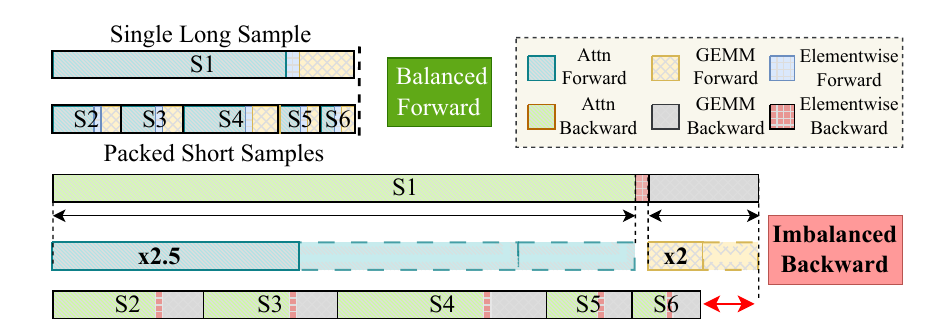}
    \caption{Imbalanced backward pass from a balanced forward pass. Despite a balanced forward pass achieved by packing shorter sequences (S2--S6) with a long sequence (S1), the backward pass exhibits significant imbalance. This is due to the higher computational cost of attention ($\times 2.5$) and GEMM ($\times 2$) operations during backward pass.}
    \label{fig:attn_backward_imbal}
\end{figure}

\subsubsection{Imbalance from Asymmetric Fwd/Bwd Costs}\label{subsubsec: asym imbalance}
A more subtle but equally critical challenge arises from the fact that balancing the forward pass does not guarantee a balanced backward pass.

\noindent\textbf{The Asymmetric Cost Profile.}
The computational costs of the forward and backward passes are inherently asymmetric. While memory-bound operations are balanced, compute-bound kernels are heavier in the backward pass. Specifically, GEMM operations are roughly twice as expensive in backward due to dual gradient computations for activations and weights, while attention kernels like FlashAttention~\cite{FlashAttention} are roughly $2.5\times$ more expensive because they must recompute some activations that were discarded during the forward pass. This creates a fundamental asymmetric cost for any given micro-batch, where the backward pass consistently demands a larger share of the computational resources.

\noindent\textbf{Failure of Forward-Balanced Packing.} 
This cost asymmetry has a critical implication: any sample-level packing strategy that perfectly balances the forward pass will inevitably become imbalanced during backpropagation. The initial balance achieved during the forward pass is merely a temporary veneer that shatters once gradient computation begins. As illustrated in Figure~\ref{fig:attn_backward_imbal}, a micro-batch \texttt{Pack1} containing a long sequence \texttt{S1} can be perfectly balanced with \texttt{Pack2} (containing shorter sequences \texttt{S2-S6}) during the forward pass. However, during the backward pass, the $2.5\times$ cost multiplier is applied to the already quadratically larger workload of \texttt{S1}. This causes the absolute increase in \texttt{S1}'s computational cost to far exceed the combined increase from all shorter sequences, turning \texttt{Pack1} into a severe straggler. Since the composition of a micro-batch is static and cannot change between passes, this ``hidden'' imbalance reintroduces the very inefficiency the packing was designed to solve, becoming a new, unaddressed source of pipeline bubbles and hardware idleness.

\section{Existing Solutions and Limitations}
\label{subsec:existing}

The landscape of existing solutions reflects a sustained effort to contain the inefficiencies caused by variable-length inputs. As summarized in Table~\ref{tab:comparison}, these approaches can be broadly categorized by their primary intervention, ranging from workload-aware scheduling to parallelism hot switching. However, none fully resolves the complex interplay between memory pressure, communication cost, and load imbalance.

\begin{table}[t]
    \centering
    \caption{Comparison of existing methods for variable-length LLM training. We evaluate each approach across memory efficiency, communication overhead, sampling order lossless-ness, and multi-level load balancing capabilities. A detailed discussion can be found in \S\ref{subsec:existing}. ($\good$: Effective, $\bad$: Ineffective/Unaddressed)}
    \label{tab:comparison}
    \scriptsize 
    \setlength{\tabcolsep}{3pt}  
    \begin{tabular*}{\linewidth}{@{\extracolsep{\fill}} c c | c c c | c c c @{}}
        \toprule
        \multirow{3}{*}{\textbf{Category}} & 
        \multirow{3}{*}{\textbf{Work}} & 
        \multirow{3}{*}{\makecell{\textbf{Memory}\\\textbf{Effic.}}} & 
        \multirow{3}{*}{\makecell{\textbf{Comm.}\\\textbf{Effic.}}} & 
        \multirow{3}{*}{\textbf{Lossless}} & 
        \multicolumn{3}{c}{\makecell{\textbf{Load Balancing}}} \\
        \cmidrule(l){6-8} 
        & & & & & 
       
        \textbf{DP} & 
        \textbf{PP} & 
        \makecell{\textbf{Fwd/}\\\textbf{Bwd}}  \\
        \midrule
        
        \multirow{2}{*}{\parbox{1.7cm}{\centering Workload-Aware Scheduling}}
        & WLB-LLM & $\bad$ & $\good$ & $\bad$ & $\good$ & $\good$ & $\bad$ \\
        & DynaPipe & $\bad$ & $\good$ & $\good$ & $\good$ & $\good$ & $\bad$ \\
        \midrule

        \multirow{3}{*}{\parbox{1.7cm}{\centering Parallelism Hot Switching}}
        & FlexSP & $\bad$ & $\bad$ & $\good$ & $\good$ & $\bad$ & $\bad$ \\
        & ByteScale & $\bad$ & $\bad$ & $\good$ & $\good$ & $\good$ & $\bad$ \\
        & HotSPa & $\bad$ & $\bad$ & $\good$ & $\good$ & $\good$ & $\bad$ \\
        \midrule

        \textbf{Ours Method} & \textbf{SlimPack} & $\good$ & $\good$ & $\good$ & $\good$ & $\good$ & $\good$ \\
        \bottomrule
    \end{tabular*}
\end{table}

\subsection{Sample Packing}\label{subsec:packing}

\noindent\textbf{Length-Based Strategies.} 
Length-based sample packing groups sequences of comparable or bounded maximum length to minimize padding. This strategy mitigates the issue of non-uniform computational load across samples, as the workload is positively correlated with the total sequence length. Nonetheless, due to the quadratic computational complexity of the self-attention mechanism, a significant disparity in computational load persists. 

\noindent\textbf{TFLOPs-Based Strategies.} 
An intuitive solution to workload imbalance is TFLOPs-based sample packing, which aims to equalize the theoretical floating-point operations per batch, accepting memory imbalance as a trade-off. The naive application of this method fails on real-world datasets characterized by long-tailed length distributions. The quadratic complexity of attention means that accommodating extreme length outliers forces shorter sequences to be packed into impractically long composites, violating hardware memory limits. Additionally, a key theoretical limitation arises when a single outlier's workload exceeds the capacity of a standard batch, rendering perfect balance unattainable. 

\smallskip\noindent Although both packing strategies improve hardware utilization, they operate at a coarse, sample-level granularity. By treating each sample as an indivisible unit, they cannot prevent a single long sequence from becoming a straggler, which then triggers the cascading imbalance in hybrid DP+PP systems. Furthermore, these packing strategies are suboptimal as they are oblivious to the asymmetry between forward and backward computation costs, using a single strategy for both phases. The following solutions are grounded in the TFLOPs-based packing framework. These solutions can be broadly categorized into two complementary strategies: \textbf{Parallelism Hot Switching} and \textbf{Workload-Aware Scheduling}.

\subsection{Workload-Aware Scheduling}\label{subsec:scheduling}

This class of solutions focuses on the data-preparation stage, reconstructing micro-batches to mitigate imbalance.

\noindent The FLOPs-only packing objective does not ensure phase balance or memory efficiency with ultra-long outliers. In response, \textbf{WLB-LLM}~\cite{WLB-LLM} employs an outlier-delay mechanism that departs from the standard data order, caching outliers for later processing to preserve training efficiency. While effective for load balancing, this strategy introduces a risk to statistical convergence by altering data stochasticity, presenting a fundamental trade-off between system performance and model integrity.

\noindent\textbf{DynaPipe}~\cite{dynapipe} introduces an adaptive scheduling mechanism to mitigate pipeline bubbles caused by imbalanced micro-batches. It achieves this by dynamically reordering the execution sequence of micro-batches within a single training iteration, a design that preserves model correctness. The core idea involves trading off memory for throughput by warming up with more forward passes. However, this very trade-off limits its efficacy against outliers, as it becomes constrained by hardware memory capacity and cannot fully eliminate the bubbles induced by severe computational imbalance.

\subsection{ Parallelism Hot Switching}
\label{subsec:adaptive_topologies}
This category of solutions dynamically alters the parallelism configuration or execution strategy at runtime in response to workload variations.

\noindent The core innovation of \textbf{FlexSP}~\cite{FlexSP} lies in its creation of heterogeneous Sequence Parallelism groups tailored to sample length. By allocating longer sequences to larger SP groups, it strives to balance the computational load among packed samples. To handle extreme outliers, FlexSP employs cross-node sequence parallelism to match their workload to that of other samples, albeit at the cost of significant communication overhead.  In terms of parallel strategy, a notable drawback is its lack of support for Pipeline Parallelism, restricting its applicability to broader model architectures.

\noindent\textbf{ByteScale}~\cite{ByteScale} proposes Hybrid Data Parallelism (HDP) to address the challenge of load imbalance in training scenarios that incorporate Pipeline Parallelism. The framework provides two balancing strategies---DP Balance and PP Balance---as trade-off solutions.
To handle outliers that induce significant computational skew, ByteScale increases the degree of Sequence Parallelism, a strategy similarly adopted by FlexSP. However, this approach for managing extreme cases introduces substantial communication overhead, mirroring a key limitation observed in FlexSP.

\noindent\textbf{HotSPa}~\cite{HotSPa} advances beyond adaptive data parallelism by dynamically switching the entire hybrid parallelism strategy—including model parallelism dimensions like Tensor Parallelism—at runtime. It dynamically applies distinct hybrid parallel strategies to sequence length based mini-batch partitions, supported by a graph compiler and a hot-switch planner that manages real-time parameter/gradient re-sharding across strategies from a shared model storage. Although this approach maximizes the theoretical optimization space by assigning an ideal parallel strategy to each data subset, it does so at the cost of substantial communication overhead. The frequent reconfiguration of the hybrid parallel plan, especially when it involves cross-machine Pipeline Parallelism, introduces inter-node communication costs that dramatically exceed those inherent to the methods of FlexSP and ByteScale.

In summary, as summarized in Table~\ref{tab:comparison}, no existing method holistically resolves the core challenges outlined in Section~\ref{subsec:challenges} without introducing significant trade-offs. Critically, it must be emphasized that parallelism hot switching aims to balance computational cost, not memory usage. This often results in allocating a larger sequence parallel groups to an individual long sequence that are computationally intensive but shorter in their packed form. Consequently, hot switching does not address the problem of memory usage.

\begin{figure}[H]
  \centering
  \includegraphics[
    trim=0mm 190mm 0mm 0mm,  %
    clip,
    width=\linewidth  %
  ]{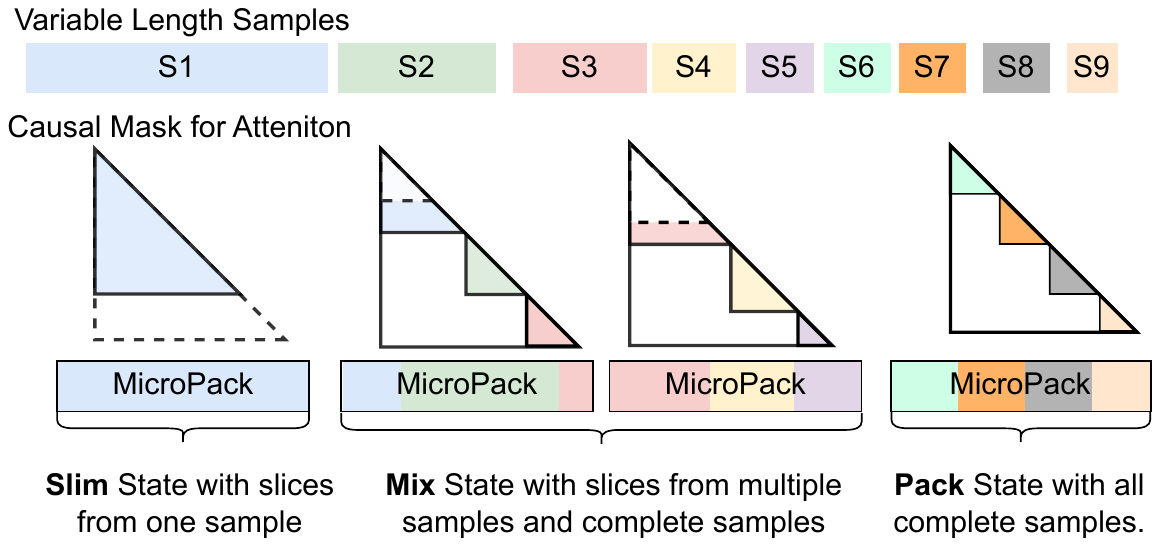}
\caption{MicroPack formation under a uniform FLOPs budget. Variable-length samples (S1–S9) are transformed into three formation states: Slim—consecutive slices from one long sample; Mix—leftover slices from a slimmed sample co-packed with complete short samples or additional slices (from other samples, order preserved) to meet the budget; Pack—all complete short samples. Note: The attention masks and slice areas are drawn solely to illustrate intra-sample slice dependencies; they are not to scale.}
  \label{fig:MicroPack}
\end{figure}

\section{SlimPack System}


\begin{figure*}[ht]
  \centering
  \includegraphics[
    trim=0mm 0mm 0mm 0mm,
    width=\linewidth
  ]{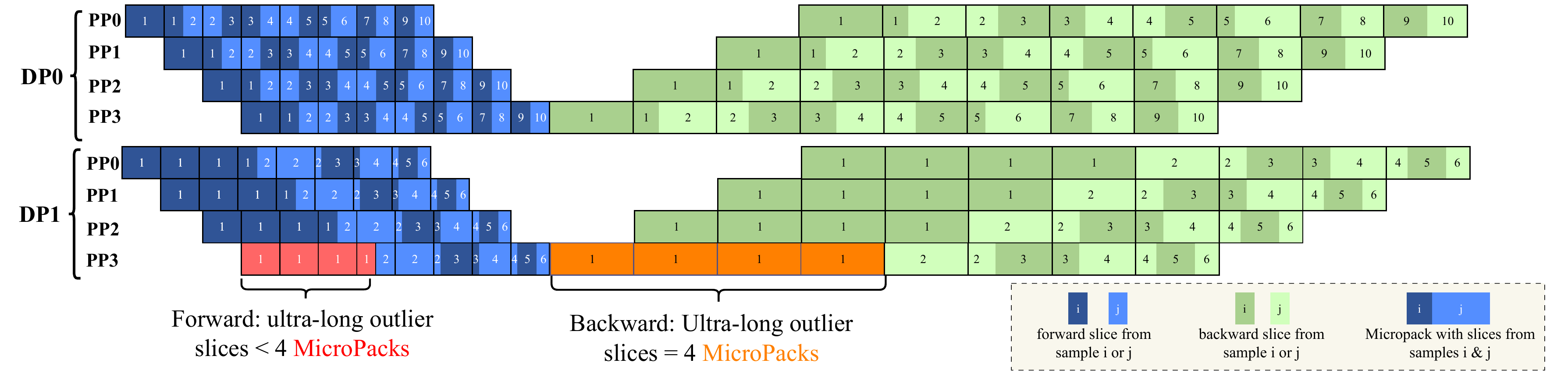}
  \caption{ SlimPack's PP schedule for 16 samples across two pipelines (10 for DP1 and 6 for DP2), each using 8 MicroPacks. During backpropagation, slices are reallocated for asymmetric partitioning to rebalance workloads. For instance, sample 1 is split differently during forward and backward passes to balance MicroPack loads. Decomposing a long sequence into slices transforms its high-variance compute cost into many low-variance chunks, ensuring each pipeline stage processes near homogeneous workloads and thus prevents straggler-induced bubbles.}
  \label{fig:slimpack-gpipe}
\end{figure*}

In this section, we present SlimPack, a zero-communication-overhead, and memory-efficient framework for training LLMs with highly heterogeneous, variable-length inputs. SlimPack introduces a principled, aggressively optimized scheduling–\\packing co-design that (i) dramatically reduces pipeline idle time, (ii) rigorously preserves slice-level packing dependencies, and (iii) substantially lowers peak activation memory.

\textbf{MicroPack} in SlimPack system serves as the minimal schedulable, budgeted container. Under a uniform FLOPs budget per MicroPack, it admits three formation states: \underline{Slim} — only consecutive slices from a single long sample; \underline{Mix} — leftover slices from a previously slimmed sample co-packed with complete short samples or additional slices (i.e., consecutive slices drawn from other samples, preserving their per-sample order) to meet the budget; and \underline{Pack} — filled entirely with complete short samples, akin to regular sample-level packing. A single sample may span multiple MicroPacks, and its slice order is always preserved as shown in Figure \ref{fig:MicroPack}. Note: a MicroPack may re-group slices during the backward pass to address forward–backward load asymmetry.

 the overall SlimPack framework comprises three MicroPack-oriented components:
 \textbf{Asymmetric Partitioning Solver} distributes global batch samples across DP groups, conditionally applies DP-Merge for extreme outliers, and determines each sample’s slim/pack state, thereby producing a family of feasible MicroPack configurations. Building on the chosen configuration, the \textbf{PackFlow Pipeline Parallelism scheduling} parameterizes the mapping and ordering of MicroPacks across stages—equalizing per-stage compute, limiting GPU memory peaks, and reusing existing DP/PP/CP collectives with zero additional communication. Given a PackFlow-instantiated schedule,
 \textbf{DAG-based Simulator} faithfully captures data dependencies, memory pressure, and pipeline bubbles, quantitatively evaluate and compare the scheduled candidates, enabling principled selection of the configuration that maximizes end-to-end efficiency under highly variable sequence lengths and finally dispatches the selected schedule for runtime execution.

\subsection{Balanced Packing \& Asymmetric Partitioning Solver}
\label{sec:solver-slice-partitioning}

We formulate LLM packing as a step-time minimization problem under per-rank memory and compute budgets. Exploiting autoregressive causality, sequences are sliced into primitives and assembled into fixed-budget MicroPacks. Because memory scales roughly linearly with sequence length while attention/compute grow superlinearly, the solver balances packing granularity against end-to-end efficiency by providing equal memory budgets across DP ranks and packing compute locally, keeping model/optimizer states stationary, and controlling pipeline bubbles via the MicroPack count rather than inflating to the global worst-case sequence length.

\subsubsection{Two-Phase Objective Formulation}
\label{sec:two-phase-solver}
Leveraging the observations from  \S\ref{sec:scaling_disparity}, we split the solving process into two phases with corresponding objectives. 
\mysubsubsection{Phase 1} Uniform per-DP capacity assignment.

We initialize a compute-balanced target by assigning each DP rank the same forward-FLOPs capacity:

\[
C^{\mathrm{flops}}_{\mathrm{dp_i}} = \tfrac{1}{DP}\sum_{x\in\mathcal{B}} f(x),\quad
f(x)=f^{\mathrm{fwd}}(x)
\]
where \(f^{\mathrm{fwd}}(x)\) denotes the forward FLOPs of sample \(x\).
Samples in the global batch are sorted by required FLOPs and then distributed across DP groups to approximate this uniform capacity.

\mysubsubsection{Phase 2} Intra-DP slim/pack strategy to reach local workload balance.

After each DP group receives its assigned samples, this DP forward‐pass budget is split across \(m\) MicroPacks, giving each MicroPack a target $\tau_{\mathrm{MicroPack}}$ of
\[
  \tau_{\mathrm{MicroPack}} =\frac{{\sum_{x\in dp_i} \mathrm{f}^{\mathrm{fwd}}(x)}}{m}, \quad
  m = i \times p,\quad
  i \in \mathbb{Z}^+.
\]

The solver apples \emph{slim} (slice-level partitioning ) to long sequences, distributing slices from the sample across multiple MicroPacks so that each pack’s workload is close to \(\tau_{\mathrm{MicroPack}}\); for short samples, we \emph{pack} (sample-level packing) multiple samples into a single MicroPack until the per-pack budget is met.

Note that each DP replica may choose a different \(m\); accordingly, the per-replica MicroPack target can vary across replicas.
 Since our workload analysis relies on forward FLOPs, we model it as a Mixed-Integer Linear Program (MILP) problem to translate FLOPs targets into sequence lengths and thereby determine the exact slice boundaries.

\begin{figure}[ht]
  \centering
  \includegraphics[
    trim=0mm 115mm 0mm 0mm,  %
    clip,
    width=\linewidth  %
  ]{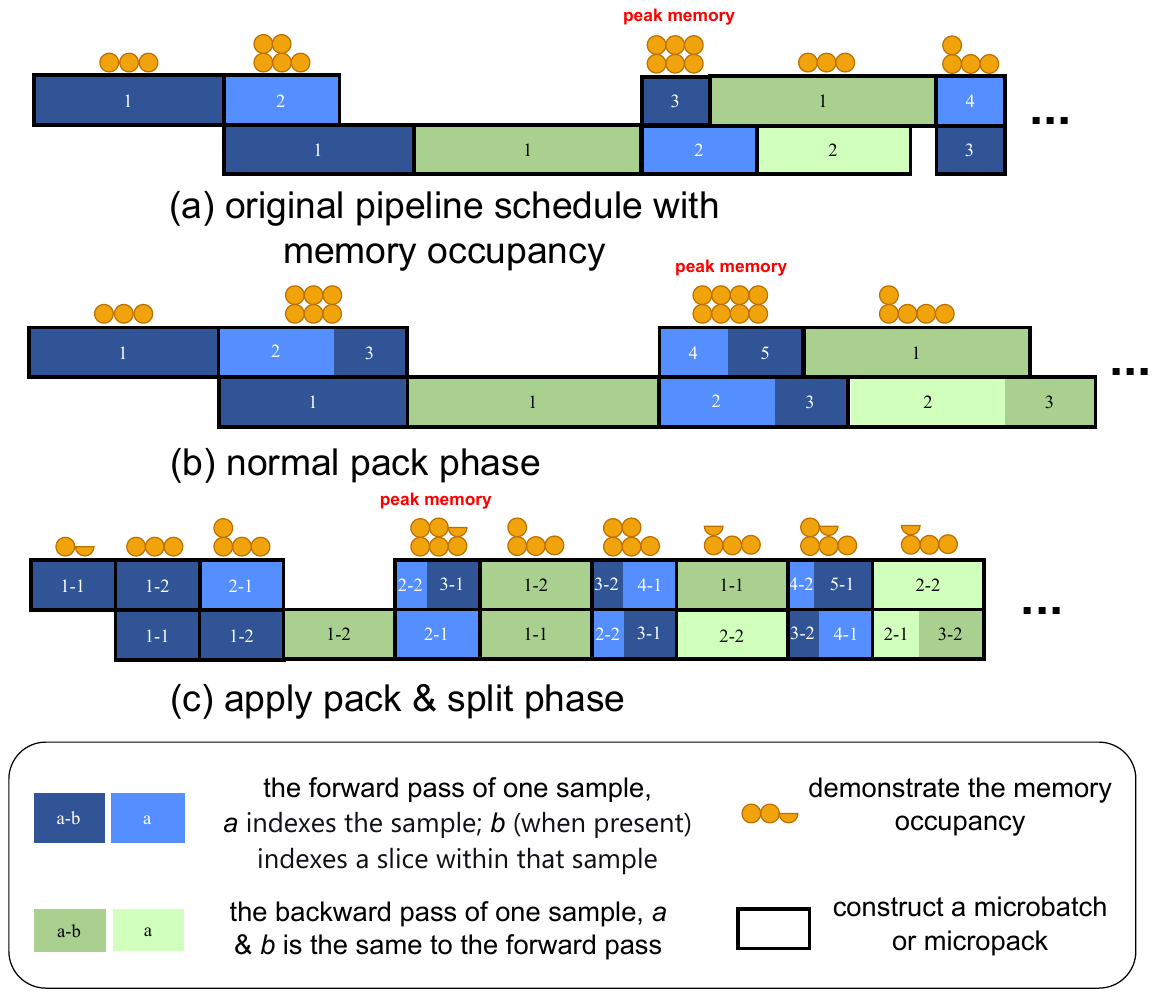}
  \caption{The illustration of the memory consumption of Microbatch/Micropack under three scheduling strategies. In the original 1F1B schedule (a), Samples 1 and 2 hold 3 and 2 memory units, whereas Samples 3-5 use only 1. An optimized schedule with a packing phase (b) reduces idle time but increases peak memory. In contrast, applying a split and pack phase (c) achieves both reduced idle time and lower peak memory consumption.}
  \label{fig:memory_advantage}
\end{figure}

\mysubsubsection{DP-Merge: On-Demand Execution Regime for Ultra-Long Samples}

While the \emph{slim} and \emph{pack} mechanisms work well for typical workloads, rare extreme long outliers \( x^\star \) with FLOPs \(f(x^\star)\gg C^{\mathrm{flops}}_{\mathrm{dp_i}}\) can behave as stragglers that inflate the system step time. Importantly, in the packing regime, the MicroPack closest to the per-GPU memory limit is not necessarily the one incurring the largest FLOPs; memory and compute outliers need not coincide. To neutralize such computation outliers, we propose a novel technique called 
\textbf{DP Merge}. When the outlier \( x^\star \) appear on \( \mathrm{dp}_i \) 
, we form a group \(G\subseteq{DP}\) with size \(|G|=g\) (\(1\le g\le DP\)). The merged ranks operate as a \emph{super-DP} and a context-parallel with  \(CP=g\) is applied as it provides the \(1/g\) per-rank scaling.
The addition context parallel dim \(g_{cp}\) spread \( x^\star \) accross rank, reducing the \emph{per-rank} costs to  
\[
f_{\mathrm{eff}}(x^\star)=\frac{f(x^\star)}{g_{\mathrm{cp}}}
\;\le\;
\min_{j\in \mathrm{DP}\setminus G} C^{\mathrm{flop}}_{\mathrm{DP_j}}.
\]
Because model and optimizer states remain local (no cross-rank state reshuffling), DP Merge is practical as an on-demand mitigation for extreme outliers and it mitigates the straggler effect, bringing the affected rank’s per-step compute into near alignment with stage-level and DP-level capacities.

\begin{figure}[ht]
  \centering
  \includegraphics[
    trim=0mm 160mm 0mm 0mm,  %
    clip,
    width=\linewidth  %
  ]{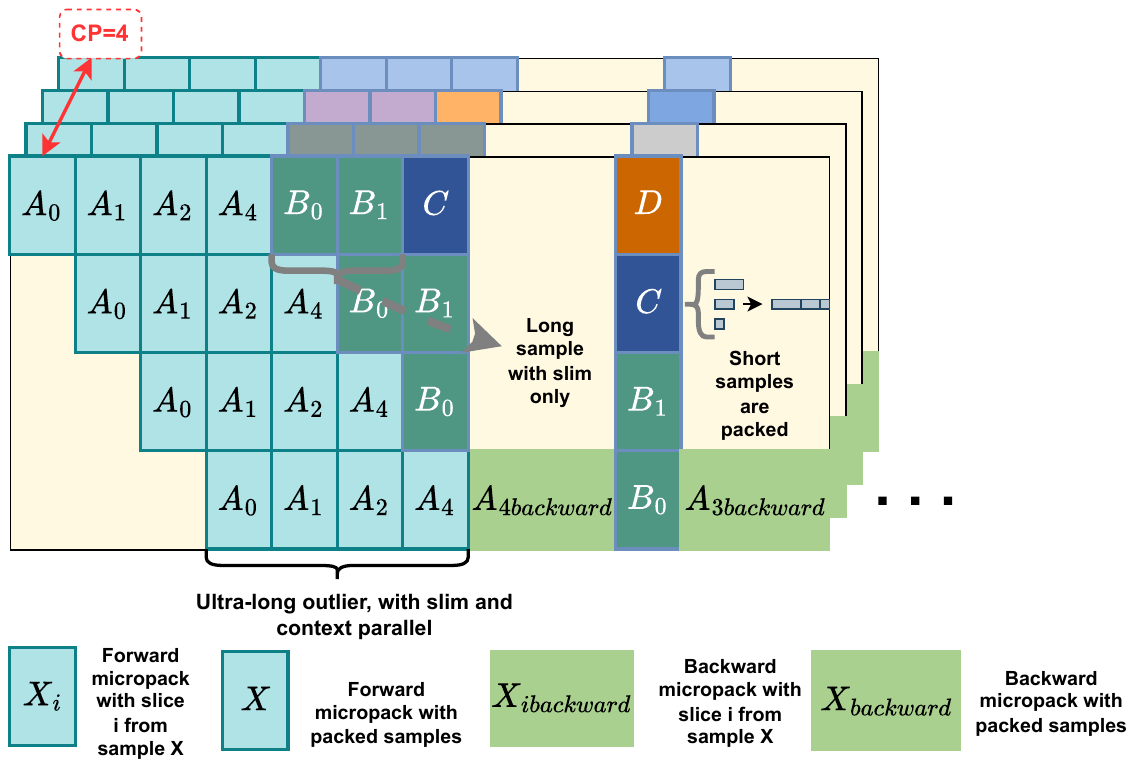}
\caption{Under DP-Merge, a super-DP with context-parallel size 4 (CP=4) slices the ultra-long outlier into MicroPacks across four devices. All other samples follow SlimPack (\emph{slim}+\emph{pack}); long sequences may be \emph{slim}-sliced into multiple MicroPacks and executed sequentially on the local device, whereas short sequences are packed into a single MicroPack as the per-pack budget permits.}
  \label{fig:dp_merge_schedule}
\end{figure}

\subsubsection{Asymmetric Partition.} \label{sec:asym-partition}
Although slice-level scheme equalizes forward-pass variance, FlashAttention’s backward recomputation obeys a steeper cost curve and reintroduces imbalance as revealed in sec~\ref{subsubsec: asym imbalance}. To address this imbalanced backward workload, we precompute each sample’s backward FLOPs and store its backward optimal slice-level partitioning strategy associated with the given number of MicroPacks without breaking the relative order of MicroPacks. At runtime, when the pipeline transitions to the backward phase, MicroPacks will dynamically apply this strategy, re-dividing slices and re-selecting samples so that each MicroPack carries an equal backward workload. The asymmetric partition and slice‐level packing scheme breaks down high‐variance cost of large samples into numerous low‐variance chunks, distributes them across balanced MicroPacks, and thus eliminates pipeline imbalance bubbles.

Finally, the solver generates multiple candidate configurations with different number of microbatches, and each of them is evaluated by a high-fidelity simulator modeling per-slice runtimes data dependency, load-imbalance bubbles, warmup or cooldown overhead, and overall GPU memory usage. The configuration with the highest estimated end-to-end throughput under the memory constraints is chosen for production-scale training runs.

\subsubsection{System-Level Analysis of SlimPack}
\mysubsubsection{Alleviating Memory Pressure}
Due to the scaling disparity of memory ($O(Lh^2)$) and computation/attention ($O(L^2h)$), naïve global packing forces short sequences to match the longest sequence’s FLOPs within a microbatch, as shown in Figure \ref{fig:memory_advantage}(b) , driving per-stage peak activations toward the global worst case. Under SlimPack system, each DP rank constrains its own activation footprint by tuning its MicroPack budget and MicroPack count.

Long samples are sliced into MicroPacks and executed sequentially, as shown in Figure \ref{fig:memory_advantage}(c) . From the pipeline perspective, this ensures that peak memory in the warm-up phase is accumulated for only a single sample at a time, rather than retaining multiple samples—including outliers—as in conventional microbatch execution.

\mysubsubsection{Eliminating DP/PP Bubble}
The primary source of DP-level imbalance is ultra-long samples that overshoot a rank’s FLOPs/memory budget and become step-time stragglers. When such an outlier is detected, DP-Merge aggregates multiple DP ranks into a temporary super-DP and applies context (Ring/Ulysses) parallelism, spreading the outlier’s sequence across \(g \) ranks so each rank’s exposed FLOPs/activations drop by \(\approx1/g \). This neutralizes the straggler and restores global DP compute balance without moving model/optimizer states. On the pipeline side, SlimPack keeps bubbles controllable along two axes: (i) warm-up/cool-down bubbles shrink as long samples are slimed into multiple MicroPacks thereby increasing the MicroPack count \(m \) and reducing the warm-up fraction, and (ii) stage-imbalance bubbles are mitigated by balanced packing/partitioning. Our asymmetric partitioning solver, introduced in \ref{sec:asym-partition}, and fixed-budget MicroPacks equalize forward/backward work across stages. In combination, DP-Merge handles outlier-induced DP skew, while SlimPack’s MicroPack granularity and asymmetric packing suppress both warm-up and imbalance bubbles.

\mysubsubsection{Reducing Communication Overhead}
Splitting a long sequence into multiple MicroPacks does not introduce extra communication beyond the baseline inter-pp-stage activation transfers. Slicing raises the number of stage-to-stage send/recv events, but the aggregate bytes are invariant; under steady-state 1F1B, the extra per-message latency is overlapped and is negligible for practical 
\(m\). When an extreme outlier triggers DP-Merge, context (Ring/Ulysses) parallelism is applied only to the outlier’s MicroPacks, while all other samples continue under the standard SlimPack schedule — introducing no additional communication and keeping the pattern at the baseline minimum as shown in Figure \ref{fig:dp_merge_schedule}.

\subsection{ PackFlow Pipeline Parallelism Scheduling}
Figure~\ref{fig:slimpack-gpipe} illustrates the overall SlimPack pipeline schedule. Both our simulator's analysis and the real run-time training adhere to this execution plan.
To facilitate clarity and understanding of the SilmPack pipeline parallel workflow, we establish the following basic conventions:
\begin{itemize}
  \item \textbf{MicroPack Execution Granularity:}  
    Each MicroPack is the smallest scheduling unit in the pipeline, and the sum of FLOPs within each MicroPack is roughly equal across all MicroPacks, ensuring load‑balanced forward and backward computation.
  \item \textbf{Intra-Pass Forward Dependency:}  
       Within the forward pass of one sample, slices must be computed in their original order to preserve the causal property of the model.
  \item \textbf{Inter-Pass Backward Dependency:}  
    The backward pass for a sample can only begin after all of its slices have completed their forward pass.
\end{itemize}

Echoing the concept of microbatches, MicroPacks traverse the entire PP workflow, covering both forward and backward passes. Each stage computes activations and sends them downstream by point-to-point communications. The solver ensures that every MicroPack carries a balanced forward workload, and the pipeline enforces strict preservation of each sample's slice order, guaranteeing correct autoregressive semantics. We employ the KV cache technique~\cite{SlimPipe, pope2023efficiently} to achieve high-throughput slice-level causal attention calculation. However, reusing the forward-pass packing during the backward pass would reintroduce imbalance as described in \ref{fig:attn_backward_imbal}. Instead, once the forward pass completes, we dynamically regroup sample slices into new backward MicroPacks according to the solver's strategy (detailed in \S\ref{sec:solver-slice-partitioning}), ensuring each MicroPack meets its backward-FLOP target.

\begin{figure*}[!tb]
  \centering
  \includegraphics[
    width=0.95\linewidth
  ]{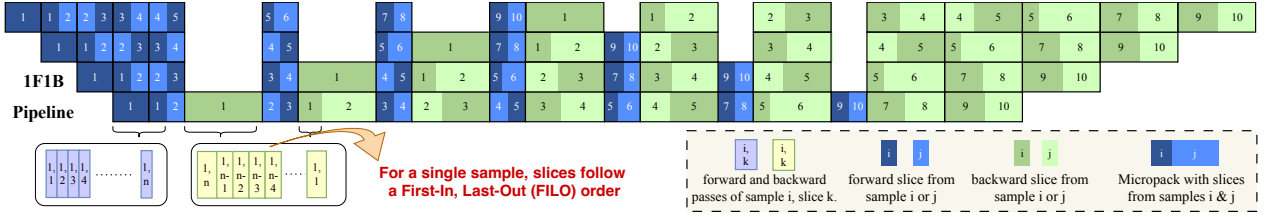}
  \caption{The 1F1B pipeline schedule of SlimPack system. Ten samples are partitioned into slices and sorted into eight MicroPacks for both forward and backward passes. MicroPacks enter the pipeline in FIFO order at the sample level, but within each sample, slices follow FILO order for correct gradient computation.}
  \label{fig:slimpack-1f1b}
\end{figure*}
\begin{figure}[h]
    \centering
      \includegraphics[
    trim=0mm 190.50mm 0mm 0mm,  %
    clip,
    width=\linewidth
  ]{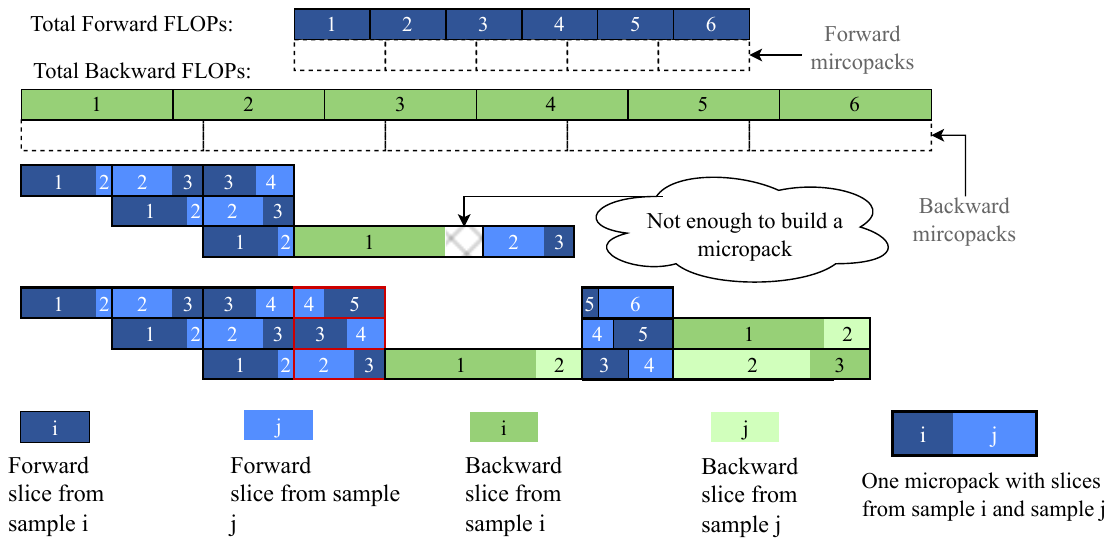}
    \caption{Pipeline schedule illustrating the injection of an extra (red) MicroPack to satisfy backward load‐balancing requirements. Six samples are divided into five forward MicroPacks and five backward MicroPacks. The backward flops from sample 1 are not enought to build one backward MicroPack and therefore, the backward of sample 1 is delayed (after forward MicroPack 2 shown in red box for last pp stage)}
    \label{fig:slimpack_1f1b_extra}
\end{figure}

The MicroPack concept, built from asymmetrically partitioned and packed slices, seamlessly integrates into the 1F1B pipeline with minimal modifications. In a standard 1F1B setup, \underline{\textit{microbatches}} are issued in FIFO (first-in, first-out) order so that each backward pass immediately follows its forward pass, eliminating activation buffering and reducing memory overhead. In SlimPack, we replace microbatches with \underline{\textit{MicroPacks}}. MicroPacks preserve FIFO order through the pipeline; nevertheless, within each sample’s slices, they follow FILO (first-in, last-out) order for correct gradient calculation, as present in Figure~\ref{fig:slimpack-1f1b}. 
Before each iteration starts, samples assigned to a data‐parallel replica are sorted in descending order of their forward‐pass FLOP cost and then sliced or grouped into MicroPacks such that, once the sample slices' \underline{\textit{backward dependency}} is met, the backward pass can launch immediately under the 1F1B protocol. However, because backward MicroPacks must satisfy specific backward-FLOP targets to guarantee balance, they can occasionally still falls short of backward-pass FLOPs requirement. In this scenario, we inject an extra forward MicroPack to increase the workload and maintain uninterrupted pipeline flow as shown in Figure ~\ref{fig:slimpack_1f1b_extra}.

\subsection{DAG-based Pipeline Simulator}

To evaluate diverse packing solutions and determine optimal pipeline scheduling strategies, we developed a sophisticated pipeline simulator as our performance model.
Traditional analytical performance models~\cite{Chimera,Alpa} often simplify calculations by neglecting inter-batch execution time variations and disparities between forward and backward pass durations, potentially yielding inaccurate estimations.
While recent simulation-based approaches employing techniques like dynamic programming~\cite{Mario} offer improved accuracy, they typically operate with a microbatch-level granularity, limiting their applicability to finer-grained pipeline schemes~\cite{li2021terapipe}.
Our DAG-based simulator addresses aforementioned issues, achieving higher fidelity in estimating both execution time and memory footprints.

\subsubsection{DAG-based pipeline representation}
We represent the entire pipeline execution as a graph $G = (V, E)$.
This representation explicitly captures task dependencies, which are crucial for accurately modeling fine-grained schedules.

\mysubsubsection{Vertices (V): tasks}
Each vertex $v \in V$ represents a computation or communication task, defined by a tuple \texttt{(model\_id, action, data\_id)}:
\begin{itemize}
    \item \texttt{model\_id}: Identifies the set of model parameters used in the task. For layer-partitioned pipeline parallelism, this typically corresponds to the pipeline stage ID (\texttt{stage\_id}) assigned to a specific device rank.
    \item \texttt{action}: Specifies the type of computation, such as Forward ($F$) or Backward ($B$). Other actions like communication, offloading and recomputation could also be incorporated as needed.
    \item \texttt{data\_id}: Uniquely identifies the data unit being processed. This can be a \texttt{microbatch\_id} or a composite identifier like \texttt{(microbatch\_id, slice\_id)} for finer granularities.
\end{itemize}
Each vertex $v$ has a weight $w(v)$ representing its estimated execution cost (time), obtained through profile-based performance modeling.

\mysubsubsection{Edges (E): dependencies}
An edge $e = (u, v) \in E$ represents a dependency, indicating that task $v$ cannot begin until task $u$ completes.
The graph must be acyclic---a cycle would imply a circular dependency, leading to a deadlock in the execution.

\begin{figure}[t]
    \centering

    \definecolor{myred}{RGB}{208, 2, 27}
    \definecolor{myblue}{RGB}{74, 144, 226}
    \definecolor{mygreen}{RGB}{126, 211, 33}
    
    \scalebox{0.85}{
    \begin{tikzpicture}[
        x=0.75pt, y=0.75pt, %
        yscale=-1, xscale=1, %
        >=Latex, %
        line width=1pt, %
        base_arrow/.style={->}, %
        red_dashed/.style={draw=myred, dashed, dash pattern=on 4.5pt off 4.5pt, base_arrow},
        blue_solid/.style={draw=myblue, solid, base_arrow},
        green_dotted/.style={draw=mygreen, dotted, dash pattern=on 1pt off 2.5pt, base_arrow},
        label_node/.style={anchor=north west, inner sep=0.75pt, align=left} %
    ]

    \draw [red_dashed] (34.64,20.36) -- (34.64,179.14);
    \draw [red_dashed] (142.38,20.36) -- (142.38,179.14);
    \draw [red_dashed] (242.94,20.36) -- (242.94,179.14);

    \draw [red_dashed] (38.23,181.14) -- (84.92,136.27);  %
    \draw [red_dashed] (149.56,181.14) -- (196.25,136.27); %
    \draw [red_dashed] (246.53,181.14) -- (296.81,136.27); %

    \draw [red_dashed] (84.92,121.31) -- (84.92,76.44);  %
    \draw [red_dashed] (196.25,121.31) -- (196.25,76.44); %
    \draw [red_dashed] (300.4,121.31) -- (300.4,76.44);   %

    \draw [blue_solid] (145.97,177.4) -- (189.07,136.27); %
    \draw [blue_solid] (60,5.4) -- (110,5.4);             %
    \draw [blue_solid] (165,9.14) -- (218.87,9.14);       %
    \draw [blue_solid] (239.35,20.36) -- (196.25,61.49);   %
    \draw [blue_solid] (60,184.88) -- (110,184.88);       %
    \draw [blue_solid] (160,65.23) -- (111.13,65.23);      %
    \draw [blue_solid] (199.84,136.27) -- (239.35,181.14); %
    \draw [blue_solid] (224.98,184.88) -- (92.1,136.27);   %

    \draw [green_dotted] (160,72.7) -- (111.13,72.7);      %
    \draw [green_dotted]  (160,128.79) -- (111.13,128.79);  %
    \draw [green_dotted] (60,12.88) -- (110,12.88);        %
    \draw [green_dotted] (60,192.36) -- (110,192.36);      %

    \draw [blue_solid] (92.1,61.49) .. controls (120.58,39.02) and (253.07,39.05) .. (293.22,61.49); %
    \draw [blue_solid] (92.1,121.31) .. controls (120.58,98.85) and (253.07,98.87) .. (293.22,121.31); %

    \node [label_node] at (1.08,1.76) {$0, F, (0, 0)$};
    \node [label_node] at (108.82,1.76) {$0, F, (0, 1)$};
    \node [label_node] at (220.36,1.76) {$0, F, 1$};
    \node [label_node] at (54.67,61.59) {$0, B, (0, 0)$};
    \node [label_node] at (54.67,121.41) {$1, B, (0, 0)$};
    \node [label_node] at (1.08,181.24) {$1, F, (0, 0)$};
    \node [label_node] at (108.82,181.24) {$1, F, (0, 1)$};
    \node [label_node] at (220.36,181.24) {$1, F, 1$};
    \node [label_node] at (162.41,61.59) {$0, B, (0, 1)$};
    \node [label_node] at (162.41,121.41) {$1, B, (0, 1)$};
    \node [label_node] at (273.95,61.59) {$0, B, 1$};
    \node [label_node] at (273.95,121.41) {$1, B, 1$};

    \end{tikzpicture}
    }

    \caption{An DAG-based representation of a 1F1B pipeline with 2 stages and 2 input microbatches, with the first microbatch divided into 2 slices. Each node represents a pass \texttt{(stage\_id, action, data\_id)}. Red dashed lines represent \textbf{inter-stage dependencies}, green dotted lines represent \textbf{inter-slice dependencies} and blue solid lines represent \textbf{schedule dependencies}.}
    \label{fig:dependency-graph}
\end{figure}
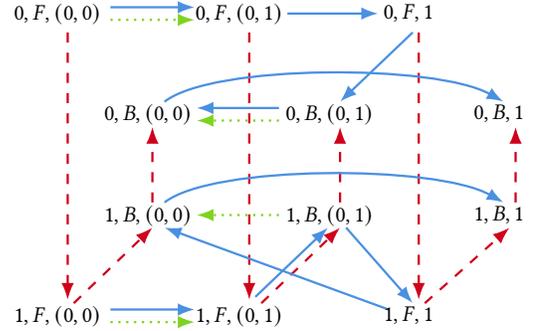

The set of edges encapsulates the ordering constraints imposed by the data flow and the pipeline schedule itself.
We construct $E$ by considering the union of the following 2 types of dependencies.

\noindent\textbf{1. Data dependencies ($E_{\text{data}}$):} These arise from the inherent data flow in LLM training.
\begin{itemize}
    \item \textbf{Inter-stage dependencies:} Data flows sequentially through stages (shown as red dashed lines in Figure~\ref{fig:dependency-graph}). For a given data unit identified by \texttt{data\_id} $id$ and a pipeline with $s$ stages, we recognize the following dependency chain:
\begin{align*}
(0, F, id) &\to (1, F, id) \to \cdots \to (s - 1, F, id) \\
&\to (s - 1, B, id) \to \cdots \to (1, B, id) \to (0, B, id)
\end{align*}
    \item \textbf{Inter-slice dependencies:} Causal attention mechanism introduces dependencies within a microbatch when sliced (shown as green dotted lines in Figure~\ref{fig:dependency-graph}). Forward pass of a slice depends on KV caches from prior slices, while its backward pass depends on KV gradients from subsequent slices.
\end{itemize}

\noindent\textbf{2. Schedule dependencies ($E_{\text{schedule}}$):} The pipeline schedule dictates the execution order of tasks assigned to a certain pipeline rank (shown as blue solid lines in Figure~\ref{fig:dependency-graph}).
These can be generated according to the scheduling strategy of the specific pipeline scheme.
Taking 1F1B schedule as an example, it first fills the pipeline with a sequence of forward passes, then enters a steady-state phase, where it interleaves the execution of a single backward pass with a forward pass. Finally, a cool-down phase drains the pipeline by executing all remaining backward passes for in-flight microbatches.

The full dependency set is $E = E_{\text{data}} \cup E_{\text{schedule}}$.
This formulation ensures the DAG accurately models all ordering constraints.

\subsubsection{DAG resolution}\label{sec:dag-resolution}
Once $G = (V, E)$ is constructed with task execution times $w(v)$ associated with each vertex $v$, we can estimate the total pipeline execution time.
This corresponds to finding the \emph{critical path} in the DAG.
We use a standard algorithm based on topological sorting and edge relaxation.

Let $Start(v)$ be the start time for task $v$, and $Finish(v)$ be its finish time.
We have:
\begin{align}
Start(v) &= \max\ab( \ab\{0\} \cup \ab\{ Finish(u) \mid (u, v) \in E \}) \\ %
Finish(v) &= Start(v) + w(v)
\end{align}
The total execution time is the maximum finish time over all tasks:
$ T_{\text{total}} = \max_{v \in V} \{ Finish(v) \} $.
Algorithm~\ref{alg:dag-solver} outlines the procedure, which runs in $\mathcal{O}(\left|V\right| + \left|E\right|)$ time.
It effectively computes the start and end time of every task in the schedule.

\begin{algorithm}[htbp]
\caption{Execution timeline calculation on the DAG}\label{alg:dag-solver}
\begin{algorithmic}[1]
\Procedure{CalculateExecutionTimeline}{$G$, $w$} %
    \State $Start(v) \gets 0$ for $v \in V$ %
    \State $topological\_order \gets$ \Call{TopologicalSort}{$G$}
    \For{$u$ in $topological\_order$}
        \State $Finish(u) \gets Start(u) + w(u)$
        \For{each $(u, v)$ originating from $u$} %
            \State $Start(v) \gets \max\ab(Start(v), Finish(u))$
        \EndFor
    \EndFor
    \State $T_{\text{total}} \gets \max_{v \in V} \ab\{ Finish(v) \}$
    \State \Return $Start$, $Finish$, $T_{\text{total}}$ %
\EndProcedure
\end{algorithmic}
\end{algorithm}

\begin{figure}[htbp]
    \centering
    \includegraphics[]{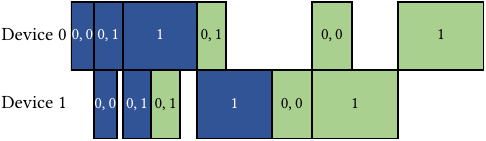}
    \caption{A pipeline schedule generated by the simulator via solving the DAG illustrated in Figure~\ref{fig:dependency-graph}. Time costs of passes are intentionally set to introduce extra bubbles.}
    \label{fig:1f1b-naive}
\end{figure}

Figure~\ref{fig:1f1b-naive} visualizes a schedule derived from solving the DAG in Figure~\ref{fig:dependency-graph}.
The task costs $w(v)$ in this example were constructed to create pipeline bubbles, which the simulator correctly identifies by computing the critical path.

\subsubsection{Memory footprint simulation}
Accurate memory estimation is crucial for identifying feasible schedules, especially under long-context scenarios. Prior models often operate at a microbatch granularity and assume a stable memory footprint during the pipeline steady phase, failing to capture the dynamics introduced by slice-level scheduling, imbalanced partitions, and techniques like activation checkpointing and offloading.

Our DAG-based simulator enables fine-grained memory tracking.
Using the execution timeline obtained previously, we simulate memory allocation and deallocation events over time.
The simulation tracks the live size of:
\begin{itemize}
    \item Model weights and optimizer states.
    \item Activations allocated during forward tasks, held until consumption by the corresponding backward tasks.
    \item KV cache and gradients managed on a per-slice basis according to the pipeline schedule.
    \item Temporary buffers used during recomputation and offloading.
\end{itemize}
The simulation proceeds chronologically.
Memory for activations is partially freed based on recomputation schedules or explicit offloading events.
This event-driven approach allows precise tracking of instantaneous memory usage, thus enabling adaptive offloading strategies as well.

\section{Implementation}
Our SlimPack is built on top of Megatron-LM~\cite{Megatron-LM}, the prevalently used parallel training framework for transfromer-based models. Computation is carried out using PyTorch as its backend.
\begin{figure}[h]
    \centering
    \includegraphics[width=\linewidth]{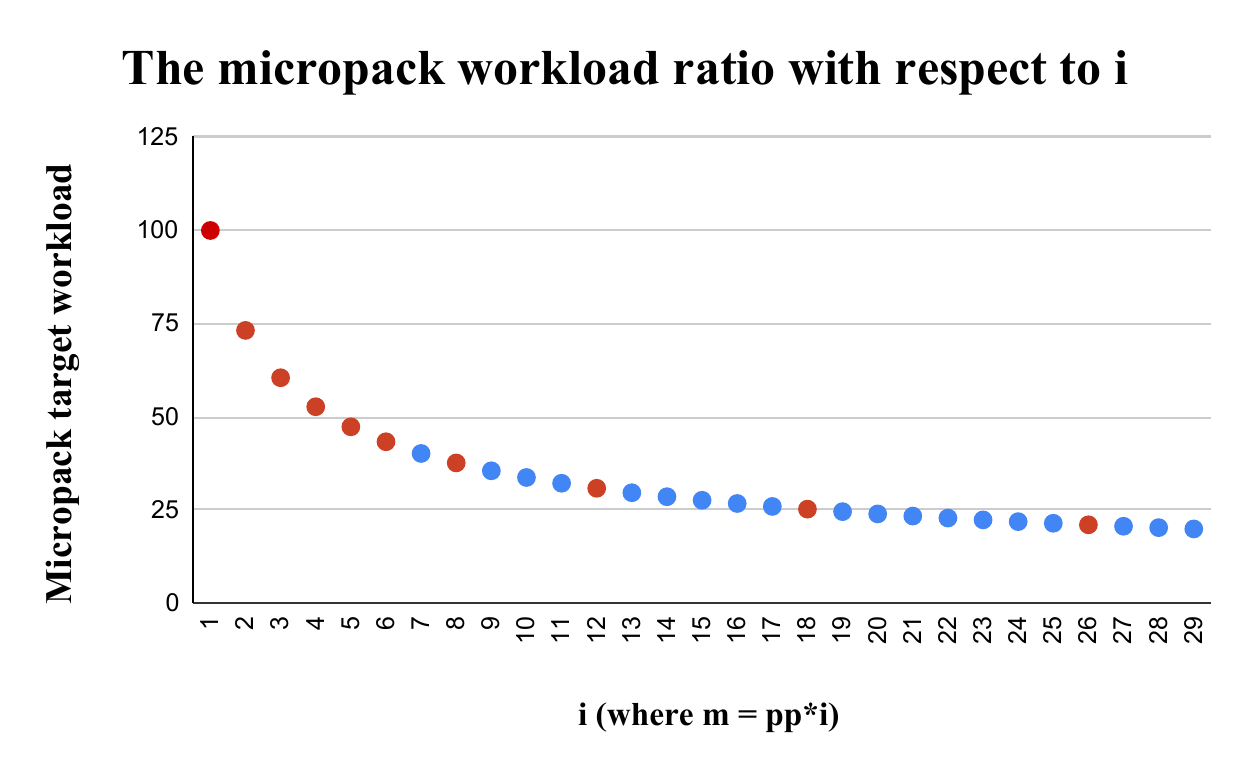}
    \caption{Given a fixed total workload \(W\), the number of micropacks \(m\) sets the per-pack budget to \(W/m\). Changing \(m\) therefore alters pack granularity and, consequently, the steady-state pipeline-bubble fraction.To prune the search space, we quantize \(m\) to multiples of the pipeline-parallel width \(\mathrm{PP}\), i.e.,\(m = i \cdot \mathrm{PP}\) with \(i \in \mathbb{Z}_{>0}\). We then evaluate only these candidates (the red dots in Figure \label{fig:bucket_size_ratio})}

    \label{fig:bucket_size_ratio}
\end{figure}
\subsection{Primitive Measurements and Search Space}

As introduced in \S\ref{sec:solver-slice-partitioning}, given the theoretical FLOPs per iteration, the SlimPack solver converts them into estimated runtime  so the simulator can estimate the end-to-end time cost. However, this conversion's accuracy hinges on each operator's hardware utilization. To compensate, we run preliminary benchmarks to characterize the peak compute‐throughput utilization for compute‐bound kernels (e.g., GEMM and Flash Attention) and profile the memory‐bandwidth utilization for memory‐bound operators (e.g., RMSNorm), then use those profiles to calibrate our runtime estimates. 

When sweeping the micropack factor 
$i$ to determine the number of micropack for a given DP workload, we observe that the budget per microbatch decreases roughly logarithmically as $i$ increases. To narrow the search space and avoid evaluating configurations with negligible impact, we sample $i$ values using a modified logarithmically spaced scheme as shown in Figure~\ref{fig:bucket_size_ratio}.

\subsection{Zero-Overhead Runtime Integration}
The SlimPack solver is integrated directly into the PyTorch data sampler, allowing the packing plan to be computed on the CPU side. To achieve low-latency and high-throughput multiple micropack assignments, the core MILP formulation and sequence‐to‐FLOPs conversion routines are implemented in \Cpp, with a lightweight thread pool. The two phase design described in \ref{sec:two-phase-solver} substantially reduces the search space. We further overlap the solver exceution with data loading and prefetching so that GPU compute stream never stalls waiting for a new packing plan. As a result, SlimPack introduces minimal overhead while delivering optimal MicroPack partitions in real time.

\section{Experiment}

\subsection{Experimental Settings}
\mysubsubsection{Enviroment}
All experiments were performed on a cluster of GPU-equipped nodes, each housing two Intel Xeon Platinum CPUs, \num{1000} GiB of RAM, and eight NVIDIA Hopper 80GB GPUs, interconnected via NVLink at 400 GB/s per GPU. For inter-node communication, every GPU is also paired with a 400 Gbps NIC. Tensor Parallelism (TP)—always combined with Sequence Parallelism (SP) are confined to individual nodes, whereas Pipeline Parallelism (PP) and Data Parallelism (DP) may span multiple nodes. Context Parallelism (CP) only crosses node boundaries when memory capacity demands it.

\mysubsubsection{Baseline} Megatron-LM~\cite{Megatron-LM} serves as our baseline framework and incorporates the same implementation of partial‐recompute and activation‐offload features as SlimPack. Both the baseline and SlimPack adopt the 1F1B pipeline schedule. We leverage parallelism (TP, DP, PP, CP) according to model scale and training setup, and enable memory optimizations (partial/full recompute, activation offload) as needed. Additionally, when handling variable-length inputs, the baseline employs the Best-Fit Packing strategy~\cite{Packing}.

\mysubsubsection{Model and Datasets}
We benchmark various LLaMA-style dense models ranging from 7B to 150B; their detailed specs appear in Table  ~\ref{tab:model_config}. GQA~\cite{GQA} is applied to the 70B and 150B models. All models are equipped with a \num{32000} sized vocabulary.
We evaluate our system on three datasets (\emph{Common Crawl}, \emph{GitHub}, and \emph{Wikipedia}) as introduced in \S\ref{sec:source_of_workload_imbalance}

\mysubsubsection{Workload and Metrics}
Three configuration axes are evaluated: context length, GPU count, and global batch size. Context length ranges from 64 K to 256 K tokens, GPU count per run scales from 128 to 256, and global batch size increases from 512 to 2048—reflecting typical large‐scale LLM training setups. We use tokens per second per GPU (TPS) as our primary performance metric and report results averaged over 20 measured iterations after a 5‐iteration warmup.

\begin{table}[t]
  \centering
  \label{tab:model_config}
      \begin{threeparttable}
    \setlength{\tabcolsep}{3pt}
    \small
\begin{tabular}{c | c | c | c | c | c}
  \toprule
  Model           & \#heads & \#groups & hidden dim & FFN dim&  \#Layers  \\
  \midrule
  Llama 7B        & 32   & --  & 4096  & 11008  & 32         \\
  Llama 13B       & 40   & --  & 5120  & 13824  & 40          \\
  Llama 70B       & 64   & 8   & 8192  & 28672  & 80           \\
  Llama 150B      & 96   & 8   & 12288 & 32768  & 96          \\
  \bottomrule
\end{tabular}
  \end{threeparttable}
    \vspace{6pt}
  \caption{Model specifications.}
  \label{tab:model_config}
\end{table}

\begin{figure*}[ht]
    \centering
    \includegraphics[]{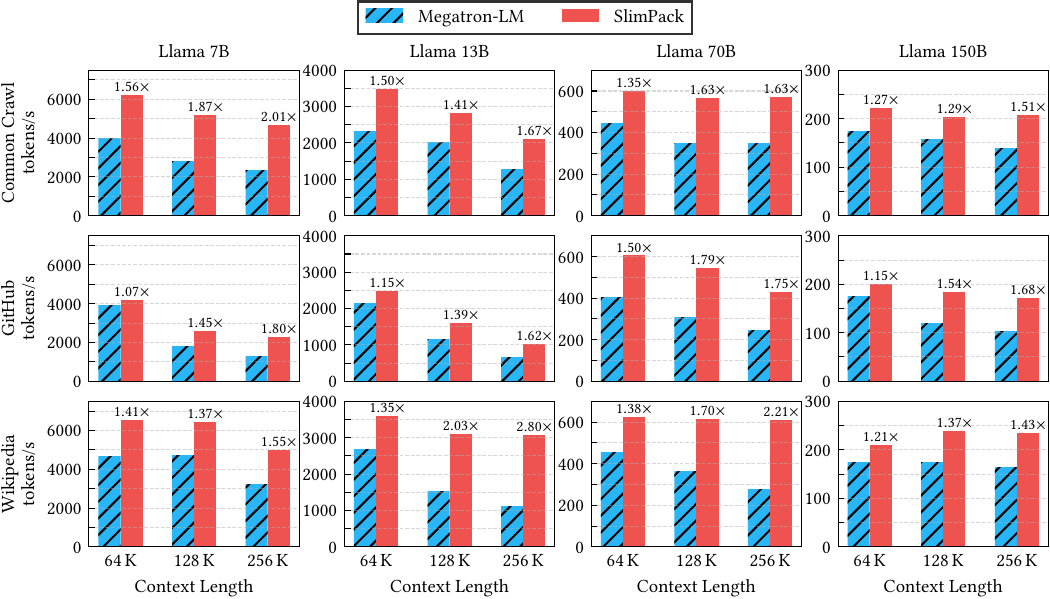}
    \caption{End-to-end performance comparison between Megatron-LM and SlimPack across different models and datasets. Annotations above the bars show the relative speedup of SlimPipe over Megatron-LM.}
    \label{fig:system}
\end{figure*}

\subsection{End-to-End Performance}

We conduct extensive experiments to evaluate the efficiency of SlimPack across three datasets, four model configurations, and three context lengths (see Table~\ref{tab:model_config}). The combined results are plotted in Figure~\ref{fig:system}. Specifically, we benchmark Llama models at different scales: Llama 7B and Llama 13B are trained on 128 devices, while Llama 70B and Llama 150B are trained on 256 devices.

SlimPack demonstrates significant efficiency improvements across all tested sequence lengths, datasets, and model sizes. Notably, for Llama 13B at a context length of 256K, SlimPack achieves a $2.8\times$ throughput improvement over the baseline. Furthermore, the performance gains of SlimPack scale progressively with longer sequences. For instance, in the Llama 150B case, SlimPack delivers a $1.15\times$ throughput improvement at 64K context length. This advantage increases to $1.54\times$ at 128K and further to $1.68\times$ at 256K, highlighting SlimPack’s superior scalability for long-context workloads. Our experiments reveal an important distinction in the scaling behavior of smaller models (Llama 7B and Llama 13B) on the GitHub dataset. While longer context lengths lead to decreased tokens per second per GPU, this stems from fundamental computational properties rather than systemic inefficiencies, such as Model FLOPS Utilization (MFU) dropping:
\begin{enumerate}
    \item Higher Attention Overhead: Attention layers dominate computation in smaller models , causing quadratic complexity to disproportionately impact total FLOPs at longer sequences. For example, at identical context lengths, attention operations consume $3.0\times$ greater share of total FLOPs in Llama 7B compared to Llama 150B.
    \item Dataset Characteristics: The GitHub corpus exhibits longer average sequence lengths than other two datasets. As evidenced by Figure~\ref{fig:data-distribution}, the GitHub dataset exhibits a right-shifted length distribution compared to Common Crawl and Wikipedia benchmarks, indicating systematically longer sequence lengths.
\end{enumerate}
In a nutshell, across all setups, SlimPack delivers throughput gains to up to $2.8\times$ over the baseline.

\subsection{Workload Balance Studies}
{\setlength{\parskip}{3pt}
\mysubsubsection{Difference in Datasets}
The three violin plots highlight how SlimPack’s asymmetric slice-level packing adapts to different data distributions. On the GitHub dataset, which displays a relatively moderate spread of sequence lengths, baseline's stragegy already shows some clustering but still suffers from occasional outliers; SlimPack compresses this spread into a tight band, virtually eliminating the ourlier and balanced the overall workload. For the long-tailed CommonCrawl data, the benefits are even more pronounced: sample-level packing produces a heavy tail of slow microbatches, whereas slice-level packing confines nearly all latencies within a narrow range, eradicating bottlenecks. Finally, on the Wikipedia distribution, where short and long sequences create two distinct performance clusters under sample-level packing, our system merges them nearly into a single, uniform mode, demonstrating its ability to smooth out both mild and extreme length variability.

\begin{figure}[t]
    \centering
    \includegraphics[]{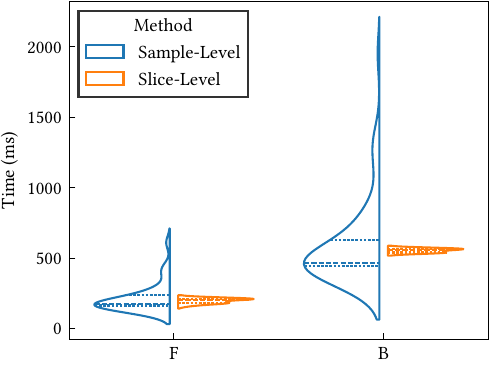}
    \caption{Violin plot comparing computation times for forward (F) and backward (B) passes between a sample-level packing method~\cite{Packing} and our proposed slice-level packing strategy. Execution times are measured on a specific device at the same iteration of two separate runs on the GitHub dataset. Each density function is normalized so the violins have equal width, with inner lines indicating quartiles.}
    \label{fig:computation_time_comparison_github}
\end{figure}

\begin{figure}[h]
    \centering
    \includegraphics[width=0.85\linewidth]{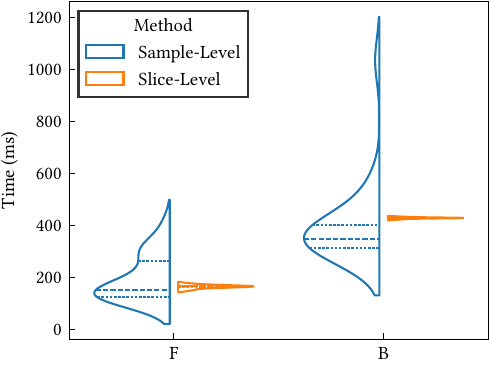}
    \caption{Comparison of forward (F) and backward (B) pass computation times between sample-level and slice-level packing on the Common Crawl dataset. The presentation style and annotations are the same as in Figure~\ref{fig:computation_time_comparison_github}.}
    \label{fig:computation_time_cc}
\end{figure}
\begin{figure}[h]
    \centering
    \includegraphics[width=0.85\linewidth]{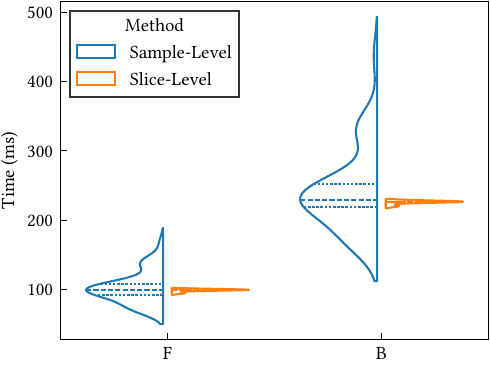}
    \caption{Comparison of forward (F) and backward (B) pass computation times between sample-level and slice-level packing on the Wikipedia dataset. The presentation style and annotations are the same as in Figure~\ref{fig:computation_time_comparison_github}.}
    \label{fig:computation_time_wiki}
\end{figure}

\mysubsubsection{Difference in MicroPack vs MicroBatch}
We evaluated the effectiveness of our proposed slice-level packing method against our sample-level packing baseline, by looking at forward and backward pass time distribution of individual microbatches or micropacks. Figure~\ref{fig:computation_time_comparison_github} presents this comparison on the GitHub dataset. The results indicate a notable advantage for slice-level packing, which exhibits a more concentrated distribution, as shown by the narrower violin plot. This implys a consistent latency across micropacks, which helps reducing PP imbalance bubbles.

We extended our analysis to Common Crawl and Wikipedia datasets, with results detailed in Figure~\ref{fig:computation_time_cc} and Figure~\ref{fig:computation_time_wiki}. Both results reaffirm the trends observed on the GitHub dataset. Across these diverse datasets, slice-level packing consistently outperforms sample-level packing, showing reduced variance in execution times. The backward pass timings are also highly concatenated, thanks to the asymmetric partitioning scheme.

Collectively, the data presented suggests that slice-level packing offers considerable improvements in terms of workload balance over traditional sample-level packing. This improved efficiency can translate to smaller pipeline imbalance bubbles, particularly for datasets with long-tailed distribution.
}
\subsection{Simulator Accuracy}

\begin{figure}[h]
    \centering
    \includegraphics[width=0.85\linewidth]{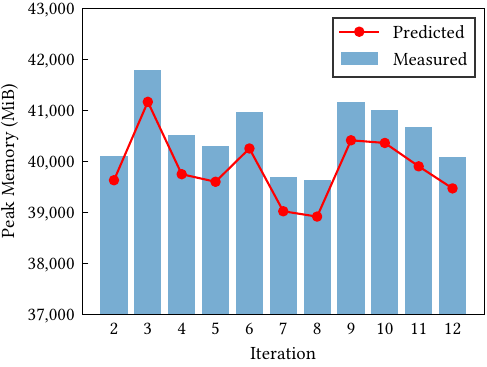}
    \caption{Comparison of measured (bars) and predicted (markers) peak GPU memory consumption in $\unit{\mebi\byte}$ from iterations 2 to 12.}
    \label{fig:memory_diff}
\end{figure}

To evaluate the precision of our memory model, we conducted an experiment using the Llama~13B model. The experiment were run on the \emph{GitHub} dataset with 16 NVIDIA Hopper 80GB GPUs. \texttt{torch.cuda.max\_memory\_allocated} is used to measure peak GPU memory usage at the end of each iteration.

As illustrated in Figure~\ref{fig:memory_diff}, our predictions align well with the actual peak GPU memory consumption.
The Mean Absolute Percentage Error (MAPE) between the predicted and measured values is merely 1.6\%.
The predicted values are consistently lower than the measured peak memory.
This underestimation is likely introduced by factors not explicitly captured in the model (e.g., communication buffers).
The systematic under-prediction does not adversely affect our packing resolution because the memory model is primarily used as a conservative filter to discard packing strategies that would lead to out-of-memory (OOM) errors.

\section{Conclusion}

In this paper, we observe critical inefficiencies in variable-length LLM training, namely the cascading imbalance in hybrid parallel systems and workload skew from asymmetric forward-backward costs, which conventional packing strategies fail to resolve. Therefore, we introduce \textbf{SlimPack}, a optimized framework that reimagines data packing by decomposing samples into fine-grained slices. Its core innovation, Asymmetric Partitioning, creates balanced scheduling units (MicroPacks) uniquely tailored for the different demands of the forward and backward passes. Guided by a two-phase solver and a high-fidelity simulator, SlimPack holistically resolves imbalances across all parallel dimensions with minimal communication overhead. Extensive experiments show SlimPack achieves up to a $2.8\times$ training throughput improvement over strong baselines under 256K context length, presenting a robust and scalable solution that breaks the conventional trade-off between workload balance and resource efficiency for long-context model training.


\bibliographystyle{ACM-Reference-Format}
\bibliography{reference}

\end{document}